\documentclass[letterpaper]{article} 
\usepackage[preprint]{aaai2027}
\usepackage[hyphens]{url} 
\usepackage{graphicx} 
\usepackage{natbib} 
\usepackage{bibunits}
\usepackage{caption} 
\usepackage{booktabs}
\usepackage{multirow}
\usepackage{amsmath}
\usepackage{amsfonts}

\usepackage[ruled,linesnumbered]{algorithm2e}

\SetAlgoCaptionSeparator{\enspace}
\DontPrintSemicolon
\SetInd{0.8em}{1.6em}
\SetNlSty{scriptsize}{}{:}
\newcommand{\algphase}[1]{%
  \vspace{0.15em}\par
  \noindent\textbf{#1}\par
  \vspace{0.15em}
}

\definecolor{rpmblue}{HTML}{315E86}
\definecolor{rpmlight}{HTML}{F3F7FA}

\title{$D^{2}R^{2}$: Discrete Diffusion with Regulation Reinforcement for Single-Cell Perturbation Prediction}

\author{
Ninghan~Fan\textsuperscript{\rm 1}\equalcontrib,
Qi~Liu\textsuperscript{\rm 1}\equalcontrib,
Xunuo~Zhu\textsuperscript{\rm 1}\equalcontrib,
Yukai~Sun\textsuperscript{\rm 1}\equalcontrib,
Luyuan~Chen\textsuperscript{\rm 1},
Xuheng~Zhou\textsuperscript{\rm 1},\\
Yuetian~Du\textsuperscript{\rm 1},
Ming~Kong\textsuperscript{\rm 1},
Xiaojun~Zhu\textsuperscript{\rm 1},
Jie~Liu\textsuperscript{\rm 2}\corresponding,
Zhan~Zhou\textsuperscript{\rm 1}\corresponding,
and Qiang~Zhu\textsuperscript{\rm 1}\corresponding
}
\affiliations{
\textsuperscript{\rm 1}Zhejiang University \qquad
\textsuperscript{\rm 2}City University of Hong Kong\\
fanninghan@zju.edu.cn, jliu.ee@my.cityu.edu.hk,\\
zhanzhou@zju.edu.cn, zhuq@zju.edu.cn
}

\begin{document}

\maketitle

\begin{bibunit}[aaai2027]
\begin{abstract}
Predicting single-cell transcriptomic responses to genetic perturbations is central to functional genomics and virtual-cell modeling. Existing approaches, however, typically predict an entire expression profile as a whole, leaving the order in which individual gene responses are generated unmodeled. To address this problem, we introduce \textbf{$D^{2}R^{2}$} (\textbf{D}iscrete \textbf{D}iffusion with \textbf{R}egulation \textbf{R}einforcement), which reformulates perturbation prediction as regulation-guided gene-wise progressive generation. A Masked Discrete Diffusion Model represents expression as ordinal tokens and reconstructs a fully masked profile step by step, allowing generated gene responses to condition those that remain masked. A Regulatory Policy Module initializes the generation policy from a gene regulatory network inferred from control cells and adapts it to the perturbation and current partially generated state. Then, group-relative policy optimization refines only the ordering policy using final perturbation-effect agreement as reward. Across Norman19 and VCC-H1, $D^{2}R^{2}$ achieves the best performance on all five metrics on Norman19 and remains competitive on H1. Controlled ablations holding the generator and generation budget fixed show that biological-prior ordering improves over random ordering and is more reliable than uncertainty-based heuristics, whereas reversing the biological-prior ordering degrades every metric. Biological analyses further show that the refined policy prioritizes regulatory genes early while promoting perturbation-specific transcription factors and responsive genes. These results establish gene generation order as an effective, controllable, and biologically interpretable dimension of single-cell perturbation prediction.

\end{abstract}

\section{Introduction}

\begin{figure}[t]
    \centering
    \includegraphics[width=\columnwidth]{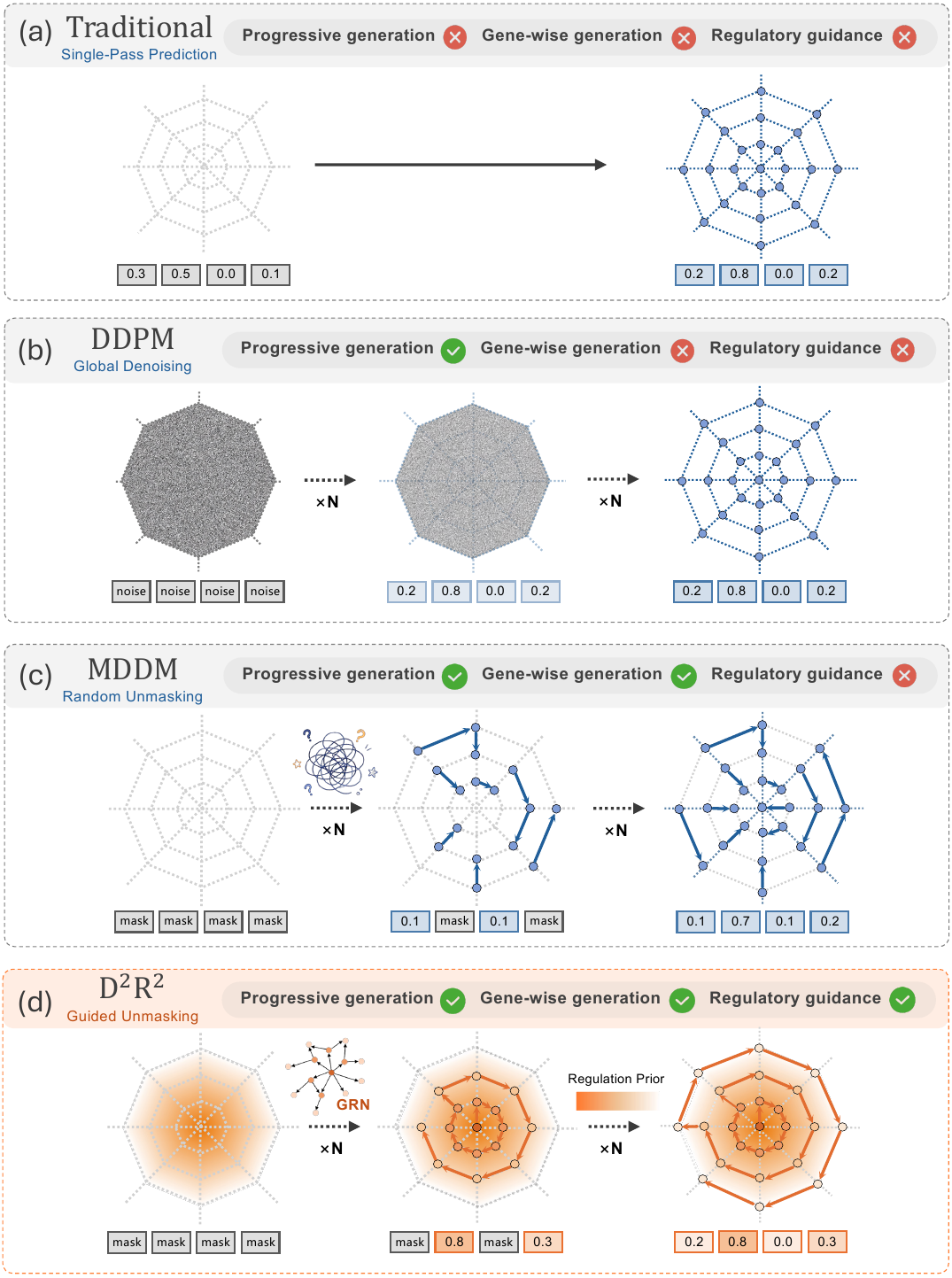}
    \caption{\textbf{Paradigm comparison.} Each row summarizes paradigm properties and traces gene-expression generation, with web nodes and aligned vector entries denoting genes and their expression values. We compare traditional single-pass prediction, DDPM-based global denoising, and two forms of $D^{2}R^{2}$: random unmasking (MDDM) and regulation-guided unmasking.}
    \label{fig:overview}
\end{figure}

Accurate prediction of cellular responses to perturbations is a central challenge in functional genomics and drug discovery. Despite rapid advances in high-throughput perturbational measurements~\citep{qian2025grow}, experimentally producing perturbation data remains costly and laborious. These challenges have motivated the development of virtual cell models that simulate cellular responses in silico, offering the potential to advance biological understanding and accelerate personalized medicine~\citep{bunne2024build, roohani2025virtual}. However, accurately modeling perturbation-induced transcriptomic responses remains difficult due to complex gene regulatory mechanisms, biological heterogeneity, and dynamic cellular responses.

Methods for perturbation prediction have evolved from single-pass predictors built on transformers~\citep{adduri2025predicting,cui2024geneformer} or variational autoencoders~\citep{bereket2023modelling} to diffusion-based generative models~\citep{he2026squidiff,klein2025cellflow,yu2026scdfm}, as illustrated in Fig.~\ref{fig:overview}a and b. Despite their architectural diversity, these approaches predominantly follow a expression profile-level generation paradigm: single-pass models generate the full profile at once, whereas diffusion models iteratively denoise the entire profile. However, in both cases, the gene-wise generation order remains unmodeled. This omission matters because gene responses are not independent, predicting gene responses in a structured order would allow earlier predictions to provide informative context for later ones~\citep{freimer2022systematic,song2025decoding}. This motivates a new perturbation-prediction paradigm that explicitly models gene-wise resolution order, allowing individual gene responses to be generated progressively.

Decomposing profile generation into gene-wise generation introduces a new challenge: in what order should gene responses be generated? Simple strategies such as random selection, predictive confidence~\citep{chang2022maskgit}, or entropy~\citep{settles2009active} use either no structural information or only the generator’s instantaneous uncertainty. A biologically grounded ordering could instead exploit gene regulatory structure to identify genes likely to influence the responses of others. Yet any fixed regulatory prior derived from unperturbed control cells would capture cell-type-level structure rather than perturbation-specific dependencies. The same ordering could not determine which dependencies are most useful for a particular perturbation in the same cell line. Addressing these limitations requires an ordering mechanism that is biologically grounded at initialization yet able to adapt to both the perturbation and the evolving generation state.

To overcome the aforementioned challenges, we reformulate single-cell perturbation prediction as an gene-wise progressive generation. Based on this formulation, we propose \textbf{$D^{2}R^{2}$}, a \textbf{D}iscrete \textbf{D}iffusion model with \textbf{R}egulation \textbf{R}einforcement, which couples a Masked Discrete Diffusion Model (MDDM) for progressive expression generation with a Regulatory Policy Module (RPM) for biologically grounded, adaptive gene ordering, as shown in Fig.~\ref{fig:overview}c and d. Specifically, MDDM represents expression values as ordered discrete tokens and begins generation from a fully masked profile. At each step, RPM selects a subset of unresolved genes, and MDDM predicts their expression tokens. Once generated, these tokens remain fixed and provide explicit context for subsequent steps. RPM determines which genes to generate next by transforming a control-derived regulatory prior into a state- and perturbation-specific generation policy. Training proceeds in two stages. We first randomly mask expression tokens at sampled diffusion times and train MDDM with an $x_0$-prediction objective to reconstruct the original tokens from partially masked profiles. We then freeze MDDM, initialize RPM to reproduce an ordering prior derived from a gene regulatory network inferred from unperturbed control cells, and refine only the ordering policy with group-relative policy optimization (GRPO), using final perturbation-response agreement as the reward.

Across Norman19 and VCC-H1, experiments validate the effectiveness of $D^{2}R^{2}$. Controlled ablations with a fixed MDDM show that generation order is consequential: the biological prior outperforms random and uncertainty-based strategies, whereas reversing it consistently degrades prediction. RPM further adapts this prior to the perturbation and evolving generation state, while biological analyses show that the refined orders preserve early regulatory-gene prioritization and promote perturbation-specific transcription factors (TFs) and responsive genes. Together, these results establish generation order as an effective and biologically interpretable component of perturbation prediction.

Our contributions are threefold:
\begin{itemize}
    \item We formulate single-cell perturbation prediction as gene-wise progressive generation with MDDM, making the generation order explicit and allowing already-resolved gene responses to condition those that remain masked.
    \item We introduce RPM, which initializes an ordering policy from a control-derived biological prior and refines it with GRPO while keeping the generator fixed, producing state- and perturbation-specific generation orders.
    \item We conduct a controlled study of random, uncertainty-based, biological, reversed, and refined orders, establishing generation order and its directionality as substantive design choices.
\end{itemize}

\section{Related Work}
\subsubsection{Single-cell perturbation prediction and generative modeling.}
Single-cell perturbation prediction has been studied under diverse predictive and generative formulations. Latent-variable methods such as scGen model perturbation effects as shifts in a shared cellular space~\citep{lotfollahi2019scgen}, while disentanglement-based predictors separate cellular state, perturbation, and covariate factors to improve generalization~\citep{piran2024disentanglement}. Distributional and flow-based approaches instead learn mappings between control and perturbed cell populations~\citep{bunne2023learning,klein2025cellflow,yu2026scdfm}. More recently, foundation models use large-scale single-cell pretraining to obtain transferable representations for perturbation prediction~\citep{cui2024scgpt,theodoris2023transfer,adduri2025predicting}. Although these approaches differ in their representations and predictive targets, they produce the response through a forward prediction, latent transformation, or distributional map. Diffusion-based models introduce iterative generation by denoising cellular states or learning bridges between control and perturbed populations~\citep{tang2023general,he2026squidiff}. Their trajectories, however, are typically governed by noise schedules over the whole transcriptomic state rather than an explicit gene-wise resolution order. $D^{2}R^{2}$ treats perturbation prediction as masked discrete generation over expression tokens, making this order a controllable part of the model.

\subsubsection{Biological priors and regulatory dependencies.}
Cellular responses to perturbations are constrained by biological structure, including gene regulatory networks (GRN), transcription factor-target relationships, and pathway interactions~\citep{dixit2016perturb,norman2019exploring,freimer2022systematic,shu2021modeling,yuan2021cellbox}. Prior work incorporates such information through graph neural networks, message passing, structure-aware embeddings, or adjacency-based regularization~\citep{adduri2025predicting,theodoris2023transfer}; graph-based perturbation models similarly use gene--gene relationships to improve response prediction~\citep{roohani2024predicting}. These approaches demonstrate the value of regulatory information, but primarily use it to shape gene representations or information exchange within the predictor. Regulatory structure therefore affects how genes interact during prediction, without defining a measurable sequence in which their responses are resolved. $D^{2}R^{2}$ applies the structure at this different point: it uses a control-derived GRN to initialize the order in which gene responses are committed, after which the ordering policy is optimized for perturbation prediction.

\subsubsection{Reinforcement learning for adaptive generation policies.}
Reinforcement learning is useful in computational biology when local decisions are difficult to supervise but complete designs can be evaluated through a delayed objective. Applications include optimizing molecular properties~\citep{olivecrona2017molecular}, constructing molecular graphs under non-differentiable objectives~\citep{you2018graph}, RNA inverse folding~\citep{runge2018learning}, and protein design~\citep{lutz2023top}. A parallel line applies RL to sequential generation itself, using sequence-level rewards or actor--critic objectives to judge early actions by their downstream consequences~\citep{ranzato2015sequence,bahdanau2016actor}. $D^{2}R^{2}$ has the same delayed-reward structure, but the policy acts on generation order rather than expression values: its actions select genes to unmask, and their value depends on how well the resulting predictions serve as context for the genes generated later. This allows RPM to optimize the ordering policy while the expression generator remains fixed.

\section{Method}
\label{sec:method}

\begin{figure*}[t]
    \centering
    \includegraphics[width=\textwidth]{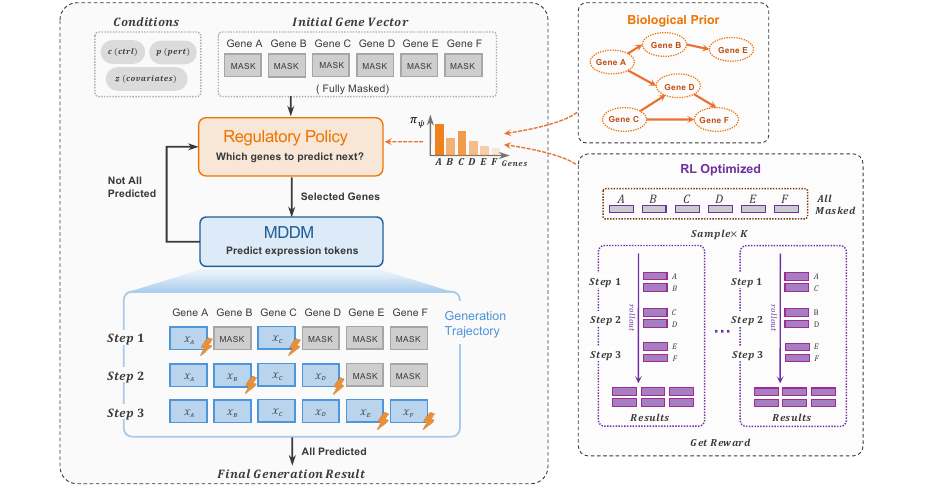}
    \caption{\textbf{Overview of the $D^{2}R^{2}$ framework.} The left panel shows the iterative generation loop: after the MDDM predicts the selected expression tokens, RPM observes the updated partial state and recomputes which genes should be resolved next. The upper-right panel shows the biological prior used to initialize the policy, while the lower-right panel illustrates GRPO that optimizes the order to each perturbation.}
    \label{fig:framework}
\end{figure*}

$D^{2}R^{2}$ first learns to generate perturbed expression tokens with a masked discrete diffusion model, and then learns which unresolved genes should be predicted at each generation step. This separates expression modeling from generation-order learning and allows us to vary the order while holding the generator fixed.

\subsection{Preliminaries}

\subsubsection{Problem formulation.}
We consider perturbation response prediction over a fixed gene set
$S=\{g_1,\ldots,g_L\}$ of $L$ genes, indexed by $\mathcal{I}=[L]=\{1,\ldots,L\}$.
For each sample, let $c\in\mathbb{R}^{L}$ denote the control expression profile, $y\in\mathbb{R}^{L}$ denote the ground-truth perturbed expression profile, $p$ denote the perturbation condition, and $z$ denote cellular covariates.
The goal is to learn a conditional generator that predicts $y$ from the conditioning tuple $u=(c,p,z)$.

\subsubsection{Discrete expression tokens.}
$D^{2}R^{2}$ represents the perturbed response in a discrete expression-token space, mapping each continuous expression value $y_i$ to an ordinal token $x_{0,i}=Q(y_i)\in\{0,\ldots,B-1\}$, where $Q(\cdot)$ is an equal-frequency binning operator with $B$ bins.
We refer to $x_0=(x_{0,1},\ldots,x_{0,L})$ as the clean discrete perturbed state, i.e., the tokenized form of $y$, where position $i$ is tied to the fixed gene identity $g_i$.
A special mask token $m$ denotes an unpredicted expression token.
The mask token makes the unresolved/resolved status of each gene explicit, thereby allowing the gene generation order to be directly specified and optimized.

\subsection{Masked Discrete Diffusion Model}

$D^{2}R^{2}$ uses a masked discrete diffusion model as the token-level generator~\citep{austin2021structured,nie2025scaling,nie2025large}, naturally allowing already generated genes to serve as explicit context for masked genes.
It learns the conditional distribution $p_\theta(x_0\mid u)$, where $\theta$ denotes the generator parameters.

\subsubsection{Forward masking corruption.}
During training, we construct a partially masked input $x_t$ from the ground-truth token state $x_0$. Following masked diffusion models~\citep{austin2021structured,nie2025scaling}, each expression token is independently retained with probability $\alpha_t=1-t$ and replaced with the mask token $m$ otherwise:
\begin{equation}
q_t(x_t\mid x_0)
=
\prod_{i=1}^{L}
\left[
\alpha_t\delta_{x_{0,i}}(x_{t,i})
+
(1-\alpha_t)\delta_m(x_{t,i})
\right].
\label{eq:mask_corruption}
\end{equation}
Here, $\delta_a(x)$ denotes one-hot distribution at $a$, taking value $1$ when $x=a$ and $0$ otherwise. This forward process enables continuous-time data corruption, with each gene $x_{t,i}$ at position $i$ equaling the original token $x_{0,i}$ with probability $\alpha_t$ and the mask token $m$ with probability $1-\alpha_t$. 

\subsubsection{Conditional denoising parameterization.}
The generator $f_\theta$ is a non-causal Transformer~\citep{vaswani2017attention} that predicts the original token at each masked gene position.
We encode the control profile, perturbation condition, and cellular covariates as
$H_c=\operatorname{Enc}_c(c)$, $H_p=\operatorname{Emb}_p(p)$, and $H_z=\operatorname{Enc}_z(z)$, and concatenate them into $H_{\mathrm{cond}}=[H_c;H_p;H_z]$.
For each fixed gene position $i\in\mathcal{I}$, the model input is $H_{t,i}=\operatorname{Emb}_x(x_{t,i})+\operatorname{Emb}_g(g_i)$.
For each bin $b\in\{0,\ldots,B-1\}$, the predictor outputs
\begin{equation}
    p_\theta(x_{0,i}=b\mid x_t,u)
    = \operatorname{Softmax}_b\!\left(
        f_\theta(H_t;H_{\mathrm{cond}})_i
    \right).
\label{eq:mdm_predictor}
\end{equation}

\subsubsection{Training objective.}
We train $f_\theta$ with an $x_0$-prediction objective on masked positions.
Given $x_0$, we sample $t\sim\operatorname{Unif}(0,1)$ and $x_t\sim q_t(x_t\mid x_0)$, and let $\mathcal{M}_t=\{i:x_{t,i}=m\}$ denote the masked gene positions. Writing $p_{\mathrm{mask}}(t)=1-\alpha_t$, the masked diffusion loss is
\begin{equation}
    \begin{aligned}
        &\mathcal{L}_{\mathrm{MDDM}}(\theta)
        = \mathbb{E}_{x_0,t,x_t}\!\left[\ell_t\right],\\
        &\ell_t =
        \frac{1}{p_{\mathrm{mask}}(t)}
        \sum_{i\in\mathcal{M}_t}
        -\log p_\theta(x_{0,i}\mid x_t,u).
    \end{aligned}
    \label{eq:mdm_objective}
\end{equation}
In practice, it is estimated by masked cross-entropy over $\mathcal{M}_t$~\citep{nie2025scaling,nie2025large}.

\subsubsection{Iterative denoising.}
At inference time, $D^{2}R^{2}$ starts from the fully masked state $x_1=(m,\ldots,m)$: at each step the predictor outputs token distributions for the currently masked positions, selected positions are updated with sampled tokens, and once a token is predicted it is kept fixed.
After generation, the discrete token sequence is converted into a continuous expression profile through a deterministic mapping that assigns each token the representative value associated with its bin.
The MDDM therefore determines the expression-token distribution for every unresolved gene, but which genes should be committed next iteration is still an open question, leaving space for exploitation.

\subsection{Gene Generation Order}

\subsubsection{Generation-order strategies.}
We formulate this open choice as a generation-order strategy that determines which masked genes are committed at each iteration.
An ordering strategy assigns a priority to every position in the current masked set $\mathcal{M}_k$ and selects a subset $a_k\subseteq\mathcal{M}_k$ to resolve.
This formulation accommodates simple orderings such as Random, as well as model-uncertainty strategies in Table~\ref{tab:order_ablation}: Confidence high/low ranks genes by the largest predicted token probability, while Entropy high/low ranks them by predictive entropy; ``high'' and ``low'' select the largest and smallest scores, respectively~\citep{chang2022maskgit,settles2009active}. These strategies provide useful controls, but they use either no structural information or only the generator's instantaneous uncertainty; neither represents how a perturbation response may propagate across genes.

\subsubsection{Biological-prior ordering.}
To introduce regulatory structure into the generation process, we infer a cell-type-specific GRN $\mathcal{G}_z=(S,A_z)$ from unperturbed control cells using DeepSEM~\citep{shu2021modeling}.
The weighted adjacency matrix $A_z$ represents directed regulatory interactions among genes.
We apply PageRank to this weighted graph to obtain a regulatory-influence score for each gene, which accounts for both its direct connections and its recursive influence through the GRN.
Sorting genes by decreasing PageRank score yields a fixed biological-prior order, with influential upstream regulators placed earlier.
At generation step $k$, this prior is restricted to the unresolved genes in $\mathcal{M}_k$ to define the biological supervision $q_k^{\mathrm{bio}}$ used to initialize RPM.
The resulting policy encourages regulatory genes to be generated first, so that their predicted responses provide context for genes resolved later.

\subsubsection{Perturbation-specific regulatory policy.}
We introduce the Regulatory Policy Module (RPM) to transform the static cell-type prior into a state- and perturbation-specific generation policy.
At step $k$, the partially generated state is $x_k$, the unresolved positions are $\mathcal{M}_k=\{i:x_{k,i}=m\}$, and the policy state is
$s_k=(x_k,p,z,\mathcal{M}_k)$.
RPM encodes the current state together with the perturbation and cellular context and produces an unmasking probability for every unresolved gene:
\begin{equation}
    \pi_\psi(i\mid s_k)
    =
    \frac{
        \exp(h_{\psi,i}(s_k)/\tau)
    }{
        \sum_{j\in\mathcal{M}_k}
        \exp(h_{\psi,j}(s_k)/\tau)
    },
    \qquad i\in\mathcal{M}_k ,
    \label{eq:unmask_priority}
\end{equation}
where $h_{\psi,i}(s_k)$ is the RPM priority logit and $\tau$ is a temperature.
The subset size follows a predefined schedule with $\sum_k|a_k|=L$.
Crucially, RPM recomputes the distribution after every update: once $a_k$ is filled by the MDDM, the resulting state $x_{k+1}$ becomes the context for deciding which genes to resolve next.
The architecture used to parameterize $h_\psi$ is described in the Appendix. We summarize the whole process in Algorithm \ref{alg:step-adaptive-rpm}.

\begin{algorithm}[t]
\small
\caption{Step-adaptive generation with RPM}
\label{alg:step-adaptive-rpm}

\textbf{Input:}
$u=(c,p,z)$, MDDM $p_\theta$, RPM $\pi_\psi$, and
$\{b_k\}_{k=1}^{K}$ satisfying
$\sum_{k=1}^{K} b_k=L$.\;

\textbf{Initialize:}
$x_1=(m,\ldots,m)$ and
$\mathcal{M}_1=\mathcal{I}$.\;

\textbf{for} $k=1,\ldots,K$ \textbf{do}\;

\Indp

\algphase{1. Generate gene tokens}

$\hat{x}_{k,i}
  \sim p_\theta(x_{0,i}\mid x_k,u),
  \quad i\in\mathcal{M}_k$.\;

\algphase{2. Select gene tokens}

$a_k\leftarrow
  \operatorname{TopK}_{b_k}
  \left\{
    \pi_\psi(i\mid x_k,p,z,\mathcal{M}_k):
    i\in\mathcal{M}_k
  \right\}$.\;

\algphase{3. Update}

$x_{k+1}\leftarrow x_k,
 \quad
 x_{k+1,a_k}\leftarrow\hat{x}_{k,a_k}$.\;

$\mathcal{M}_{k+1}
  \leftarrow
  \mathcal{M}_k\setminus a_k$.\;

\Indm

\textbf{end for}\;

\textbf{Return:}
bin-representative decoding of $x_{K+1}$.\;

\end{algorithm}

\subsubsection{Biological initialization and GRPO refinement.}
RPM is first initialized to reproduce the biological-prior policy.
For policy states encountered during generation, we minimize
\begin{equation}
    \mathcal{L}_{\mathrm{bio}}(\psi)
    =
    -\mathbb{E}_{s_k}
    \left[
        \sum_{i\in\mathcal{M}_k}
        q_{k,i}^{\mathrm{bio}}
        \log\pi_\psi(i\mid s_k)
    \right],
\label{eq:bio_initialization}
\end{equation}
which supplies a stable starting point rather than learning from scratch.

We then freeze the MDDM and biological prior and refine RPM with GRPO~\cite{shao2024deepseekmath}. Next, we construct $G$ branched rollouts from shared generation state $s$. In rollout $r$, RPM samples an ordered subset $a^{(r)}=(i_1,\ldots,i_b)$ without replacement, with likelihood
\begin{equation}
P_\psi(a^{(r)}\mid s)
=
\prod_{j=1}^{b}
\frac{\pi_\psi(i_j\mid s)}
{1-\sum_{\ell<j}\pi_\psi(i_\ell\mid s)}.
\label{eq:subset_policy}
\end{equation}
The sampled genes are resolved by MDDM, and the remaining trajectory is completed by the frozen reference policy $\pi_{\mathrm{ref}}$. Each rollout receives
$R^{(r)}=\operatorname{corr}(\hat{y}^{(r)}-c,y-c)$, which is normalized within the group to obtain $A^{(r)}$.

Let
$\rho^{(r)}=P_\psi(a^{(r)}\mid s)/P_{\psi_{\mathrm{old}}}(a^{(r)}\mid s)$
and
$\bar{\rho}^{(r)}=\operatorname{clip}(\rho^{(r)},1-\epsilon,1+\epsilon)$.
RPM is optimized using
\begin{equation}
\begin{aligned}
\mathcal{L}_{\mathrm{GRPO}}
={}&-\mathbb{E}_{r}
\left[
\min\left(\rho^{(r)}A^{(r)},
\bar{\rho}^{(r)}A^{(r)}\right)
\right] \\
&+\beta D_{\mathrm{KL}}
\left(\pi_\psi\,\|\,\pi_{\mathrm{ref}}\right).
\end{aligned}
\label{eq:grpo_objective}
\end{equation}
Training samples subsets without replacement, whereas inference recomputes RPM priorities at each step and applies deterministic Top-$K$ selection.

\section{Experiments}
\label{sec:experiments}

\subsection{Experimental Setup}

\subsubsection{Datasets.}
We evaluate $D^{2}R^{2}$ on Norman19 and VCC-H1(H1). Norman19 is a Perturb-seq dataset of gene overexpression perturbations in K562 cells, containing 287 perturbation conditions, including 131 combinatorial perturbations~\citep{norman2019exploring}. VCC-H1 is the H1 human embryonic stem cell benchmark from the Virtual Cell Challenge, containing approximately 400,000 single cells across 300 CRISPRi perturbation targets~\citep{roohani2025virtual}.
Norman19 uses a held-out-combination split to test generalization to unseen perturbation combinations, whereas H1 uses a held-out-cell split to evaluate large-scale response reconstruction under the observed perturbation distribution. The two benchmarks therefore cover complementary evaluation regimes.

\subsubsection{Baselines.}
Generative models form the primary comparison group: SAMS-VAE~\citep{bereket2023modelling}, Cell Flow~\citep{klein2025cellflow}, Squidiff~\citep{he2026squidiff}, and scDFM~\citep{yu2026scdfm}, covering VAE-based, flow-based, diffusion-based, and distribution-level approaches. We also compare with the perturbation-response predictor Biolord~\citep{piran2024disentanglement} and the virtual-cell model STATE~\citep{adduri2025predicting}. All methods use matched splits, preprocessing, and evaluation; implementation details are provided in the Appendix.

\subsubsection{Metrics.}

Following PerturBench~\citep{wu2024perturbench}, we evaluate differential expression using Pearson $\Delta$, Cos. LogFC, and Cos. LogFC Rank. These metrics measure expression changes relative to control cells, testing whether the model recovers the transcriptional programs activated or suppressed by a perturbation; Cos. LogFC Rank additionally measures whether responses to different perturbations remain distinguishable. For absolute expression, Cos. PCA measures centroid agreement between predicted and observed cells, testing whether the overall post-perturbation cellular state is reconstructed. Sym. KL further evaluates agreement between cell-population distributions, capturing response heterogeneity beyond the centroid. Metric details are provided in the Appendix.

\subsection{Main Results}

\begin{table*}[t]
\centering
\providecommand{\venue}[1]{\,{\scriptsize\textcolor{gray}{(#1)}}}
\caption{\textbf{Main results on Norman19 and H1.}
All methods are evaluated over five independent training runs using seeds $\{0,1,2,3,42\}$, and results are reported as mean~$\pm$~standard deviation. For $D^{2}R^{2}$, MDDM checkpoint is fixed across runs, while RPM is independently trained for each seed. Metrics are grouped by differential and absolute expression. Best and second-best results are bolded and underlined, respectively.}
\label{tab:main_results}
\scriptsize
\setlength{\tabcolsep}{3.2pt}
\renewcommand{\arraystretch}{1.12}
\newcommand{\err}[2]{#1{\tiny$\pm$#2}}
\resizebox{0.88\textwidth}{!}{%
\begin{tabular}{@{}llccccc@{}}
\toprule
& &
\multicolumn{3}{c}{\textbf{Differential Expression}}
& \multicolumn{2}{c}{\textbf{Absolute Expression}} \\
\cmidrule(lr){3-5} \cmidrule(lr){6-7}
\textbf{Dataset}
& \textbf{Method}
& Pearson $\Delta$ $\uparrow$
& Cos. LogFC $\uparrow$
& Cos. LogFC Rank $\downarrow$
& Cos. PCA $\uparrow$
& Sym. KL $\downarrow$ \\
\midrule

\multirow{7}{*}{Norman19}
& SAMS-VAE\venue{NeurIPS 2023}
& \underline{\err{0.553}{0.079}}
& \underline{\err{0.560}{0.043}}
& \underline{\err{0.028}{0.010}}
& \err{0.251}{0.042}
& \err{50.888}{4.215} \\
& Biolord\venue{Nat. Biotechnol. 2024}
& \err{0.404}{0.069}
& \err{0.388}{0.051}
& \err{0.046}{0.010}
& \err{0.035}{0.016}
& \err{0.526}{0.040} \\
& STATE\venue{NeurIPS 2025}
& \err{0.467}{0.020}
& \err{0.379}{0.020}
& \err{0.525}{0.017}
& \err{-0.009}{0.039}
& \err{0.273}{0.014} \\
& Cell Flow\venue{bioRxiv 2025}
& \err{0.489}{0.077}
& \err{0.373}{0.066}
& \err{0.329}{0.013}
& \err{0.084}{0.066}
& \underline{\err{0.207}{0.024}} \\
& Squidiff\venue{Nat. Methods 2025}
& \err{0.291}{0.001}
& \err{0.226}{0.002}
& \err{0.427}{0.007}
& \err{0.110}{0.000}
& \err{2.861}{0.019} \\
& scDFM\venue{ICLR 2026}
& \err{0.526}{0.010}
& \err{0.313}{0.015}
& \err{0.379}{0.008}
& \underline{\err{0.264}{0.032}}
& \err{0.390}{0.034} \\
\cmidrule(lr){2-7}
& $D^{2}R^{2}$\venue{Ours}
& \textbf{\err{0.706}{0.004}}
& \textbf{\err{0.724}{0.006}}
& \textbf{\err{0.024}{0.006}}
& \textbf{\err{0.703}{0.009}}
& \textbf{\err{0.148}{0.007}} \\
\midrule

\multirow{7}{*}{H1}
& SAMS-VAE\venue{NeurIPS 2023}
& \underline{\err{0.262}{0.017}}
& \underline{\err{0.231}{0.016}}
& \err{0.192}{0.028}
& \err{0.319}{0.042}
& \err{40.638}{3.679} \\
& Biolord\venue{Nat. Biotechnol. 2024}
& \err{0.201}{0.008}
& \err{0.159}{0.008}
& \underline{\err{0.063}{0.007}}
& \err{0.151}{0.010}
& \err{0.240}{0.018} \\
& STATE\venue{NeurIPS 2025}
& \err{0.253}{0.012}
& \err{0.178}{0.006}
& \err{0.095}{0.006}
& \textbf{\err{0.378}{0.029}}
& \err{0.272}{0.016} \\
& Cell Flow\venue{bioRxiv 2025}
& \err{0.042}{0.016}
& \err{0.028}{0.009}
& \err{0.493}{0.013}
& \err{0.014}{0.009}
& \textbf{\err{0.089}{0.006}} \\
& Squidiff\venue{Nat. Methods 2025}
& \err{0.046}{0.000}
& \err{0.049}{0.012}
& \err{0.368}{0.032}
& \err{-0.008}{0.005}
& \err{8.435}{0.372} \\
& scDFM\venue{ICLR 2026}
& \err{0.069}{0.004}
& \err{0.038}{0.009}
& \err{0.421}{0.029}
& \err{0.029}{0.010}
& \err{0.390}{0.025} \\
\cmidrule(lr){2-7}
& $D^{2}R^{2}$\venue{Ours}
& \textbf{\err{0.271}{0.002}}
& \textbf{\err{0.541}{0.004}}
& \textbf{\err{0.001}{0.000}}
& \underline{\err{0.358}{0.001}}
& \underline{\err{0.229}{0.002}} \\
\bottomrule
\end{tabular}
}
\end{table*}

Table~\ref{tab:main_results} shows that $D^{2}R^{2}$ ranks first across all metrics on Norman19 and leads in most metrics on H1. In particular, it leads all three differential-expression metrics on both benchmarks, demonstrating that it accurately captures perturbation-induced transcriptional changes while preserving the distinctions among responses to different perturbations. For absolute expression, $D^{2}R^{2}$ achieves the best centroid agreement on Norman19 and remains competitive on H1, indicating effective reconstruction of the overall post-perturbation cellular state. Its strong Sym. KL results further show that this agreement extends beyond the centroid to the cell-population distribution.

\begin{figure}[t]
    \centering
    \includegraphics[width=\columnwidth]{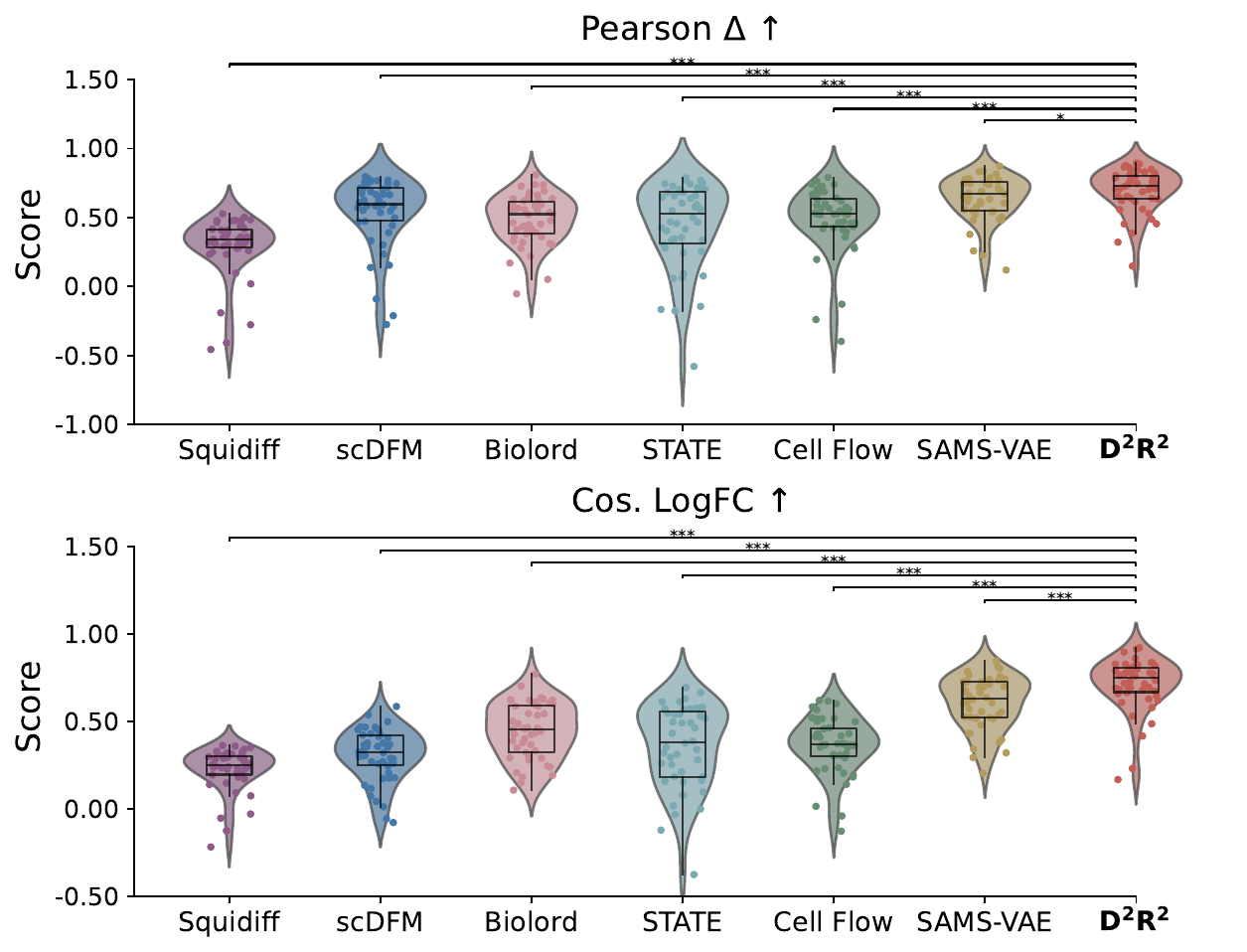}
    \caption{Violin plots showing the distribution of evaluation metrics across perturbations in Norman19. Mann-Whitney $U$ test, * P < 0.05, *** P < 0.001.
}
    \label{fig:fig3}
\end{figure}

Beyond the aggregate results, Fig.~\ref{fig:fig3} compares the per-perturbation distributions of Pearson $\Delta$ and Cos. LogFC on Norman19. $D^{2}R^{2}$ shifts both distributions toward higher response similarity, indicating that its gains are consistent across perturbations.

\begin{table*}[t]
\centering
\caption{\textbf{Generation-order ablation on Norman19.}
All methods use the same frozen MDDM.
Random averages multiple random orders.
Cells show score and relative performance gain over Random, with signs aligned so that positive always indicates improvement; for Cos. LogFC Rank and Sym. KL, this corresponds to relative error reduction. Orange intensity indicates a positive gain. Percentages are computed before rounding.}
\label{tab:order_ablation}
\definecolor{ablOrange}{RGB}{244,162,97}
\newcommand{\orderplain}[2]{%
  \makebox[7.3em][c]{#1\,{\color{black!62}(#2\%)}}}
\newcommand{\ordergain}[3]{%
  \begingroup\setlength{\fboxsep}{1.0pt}%
  \colorbox{ablOrange!#1}{\makebox[7.3em][c]{#2\,(#3\%)}}\endgroup}
\newcommand{\orderbase}[1]{%
  \begingroup\setlength{\fboxsep}{1.0pt}%
  \colorbox{gray!12}{\makebox[7.3em][c]{#1\,(ref.)}}\endgroup}
\newcommand{\ordergroup}[1]{%
  \addlinespace[3pt]
  \multicolumn{6}{c}{\bfseries\itshape #1}\\[-2pt]
  \cmidrule(lr){1-6}}
\footnotesize
\renewcommand{\arraystretch}{1.12}
\setlength{\tabcolsep}{2.4pt}
\begin{tabular}{@{}lccccc@{}}
\toprule
\textbf{Order}
& \textbf{Pearson $\Delta$} $\uparrow$
& \textbf{Cos. LogFC} $\uparrow$
& \textbf{Cos. LogFC Rank} $\downarrow$
& \textbf{Cos. PCA} $\uparrow$
& \textbf{Sym. KL} $\downarrow$\\
\midrule

Random
& \orderbase{0.623}
& \orderbase{0.720}
& \orderbase{0.032}
& \orderbase{0.597}
& \orderbase{0.183}\\

\ordergroup{Model-Uncertainty Orders}
Confidence high
& \orderplain{0.590}{$-5.3$}
& \orderplain{0.537}{$-25.4$}
& \ordergain{30}{0.030}{$+5.2$}
& \orderplain{0.353}{$-40.9$}
& \orderplain{0.410}{$-124.4$}\\

Confidence low
& \ordergain{45}{0.672}{$+7.9$}
& \orderplain{0.522}{$-27.5$}
& \orderplain{0.068}{$-114.1$}
& \ordergain{40}{0.640}{$+7.2$}
& \orderplain{0.203}{$-11.0$}\\

Entropy high
& \ordergain{10}{0.624}{$+0.2$}
& \orderplain{0.452}{$-37.2$}
& \orderplain{0.073}{$-130.9$}
& \orderplain{0.564}{$-5.6$}
& \orderplain{0.206}{$-12.5$}\\

Entropy low
& \orderplain{0.591}{$-5.1$}
& \orderplain{0.538}{$-25.3$}
& \ordergain{30}{0.030}{$+5.2$}
& \orderplain{0.355}{$-40.5$}
& \orderplain{0.399}{$-118.5$}\\

\ordergroup{Biological-Prior Orders}
Biological prior
& \ordergain{65}{0.706}{$+13.4$}
& \orderplain{0.703}{$-2.3$}
& \ordergain{70}{0.027}{$+14.4$}
& \textbf{\ordergain{75}{0.699}{$+17.0$}}
& \textbf{\ordergain{75}{0.119}{$+35.0$}}\\

Reversed prior
& \orderplain{0.573}{$-8.1$}
& \orderplain{0.533}{$-26.0$}
& \orderplain{0.064}{$-101.8$}
& \orderplain{0.509}{$-14.7$}
& \orderplain{0.197}{$-7.5$}\\

\ordergroup{Adaptive RPM Policies (GRPO)}
RPM (random init.)
& \ordergain{50}{0.677}{$+8.6$}
& \textbf{\ordergain{15}{0.734}{$+2.0$}}
& \ordergain{75}{0.026}{$+19.0$}
& \ordergain{65}{0.675}{$+13.1$}
& \orderplain{0.186}{$-1.6$}\\

RPM (bio init.)
& \textbf{\ordergain{70}{0.711}{$+14.2$}}
& \ordergain{10}{0.723}{$+0.5$}
& \textbf{\ordergain{75}{0.020}{$+35.8$}}
& \ordergain{75}{0.693}{$+16.1$}
& \ordergain{75}{0.138}{$+24.5$}\\

\bottomrule
\end{tabular}
\end{table*}

\subsection{Generation-Order Ablation}

To isolate the effect of generation order, we compare different orders. Table~\ref{tab:order_ablation} reports a controlled ablation in which every configuration uses the same frozen MDDM and generation budget; the only difference is which masked genes are selected at each step.

\subsubsection{Model uncertainty does not identify a useful generation order.}
Overall, model-uncertainty-based strategies provide little guidance for organizing gene generation. Although some uncertainty orders improve individual metrics, these gains are often accompanied by substantial deterioration elsewhere. For example, prioritizing low-confidence genes improves Pearson $\Delta$ and Cos. PCA but worsens Cos. LogFC and perturbation discriminability, suggesting that local uncertainty does not identify genes that provide useful context for subsequent generation.

\subsubsection{Our biological prior provides a strong directional generation scaffold.}
The biological-prior ordering introduced in $D^{2}R^{2}$ improves most metrics over Random, with only a modest decrease in Cos. LogFC, showing that regulatory structure alone provides an effective scaffold for gene generation. Further, reversing the prior order causes every metric to fall below Random, revealing that following regulatory dependencies in the wrong direction is actively detrimental rather than merely uninformative. Thus, the prior contributes not merely by imposing structure, but by organizing generation along a predictive regulatory direction.

\subsubsection{RPM turns static prior into perturbation-specific generation policies.}
Even with random initialization, RPM improves multiple metrics over Random, showing that response-level reinforcement learning can discover useful generation orders. With biological initialization, RPM converts the fixed prior into perturbation-specific policies that improve all three perturbation-effect metrics, with only modest trade-offs in centroid- and population-level agreement.

\subsection{Biological Analysis}

\subsubsection{RPM preserved the generation prioritization of regulatory genes.}
As shown in Fig.~\ref{fig:fig4}a, regulatory genes were strongly enriched at early generation steps and declined thereafter, whereas the proportion of regulated genes progressively increased across steps. This result suggests that RPM tends to establish upstream regulatory context before generating downstream responses.

\subsubsection{RPM prioritizes perturbation-specific TFs to earlier generation steps.}
Next, we test whether RPM captures perturbation-specific regulatory rewiring. Perturbation-specific TFs are those inactive in unperturbed cells but acquire regulatory activity after perturbation (details in the Appendix). We quantify the RPM-induced change in generation priority as $\Delta\mathrm{Order}=\mathrm{Order}_{\mathrm{Bio\text{-}prior}}-\mathrm{Order}_{\mathrm{RPM}}$, where a positive value indicates that RPM generates the gene earlier. In Norman19, perturbation-specific TFs exhibit significantly higher $\Delta\mathrm{Order}$ than non-specific TFs (Fig.~\ref{fig:fig4}b), demonstrating that RPM preferentially advances perturbation-specific regulators.

\subsubsection{RPM-promoted genes capture perturbation-specific transcription responses.}
\begin{figure}[t]
    \centering
    \includegraphics[width=\columnwidth]{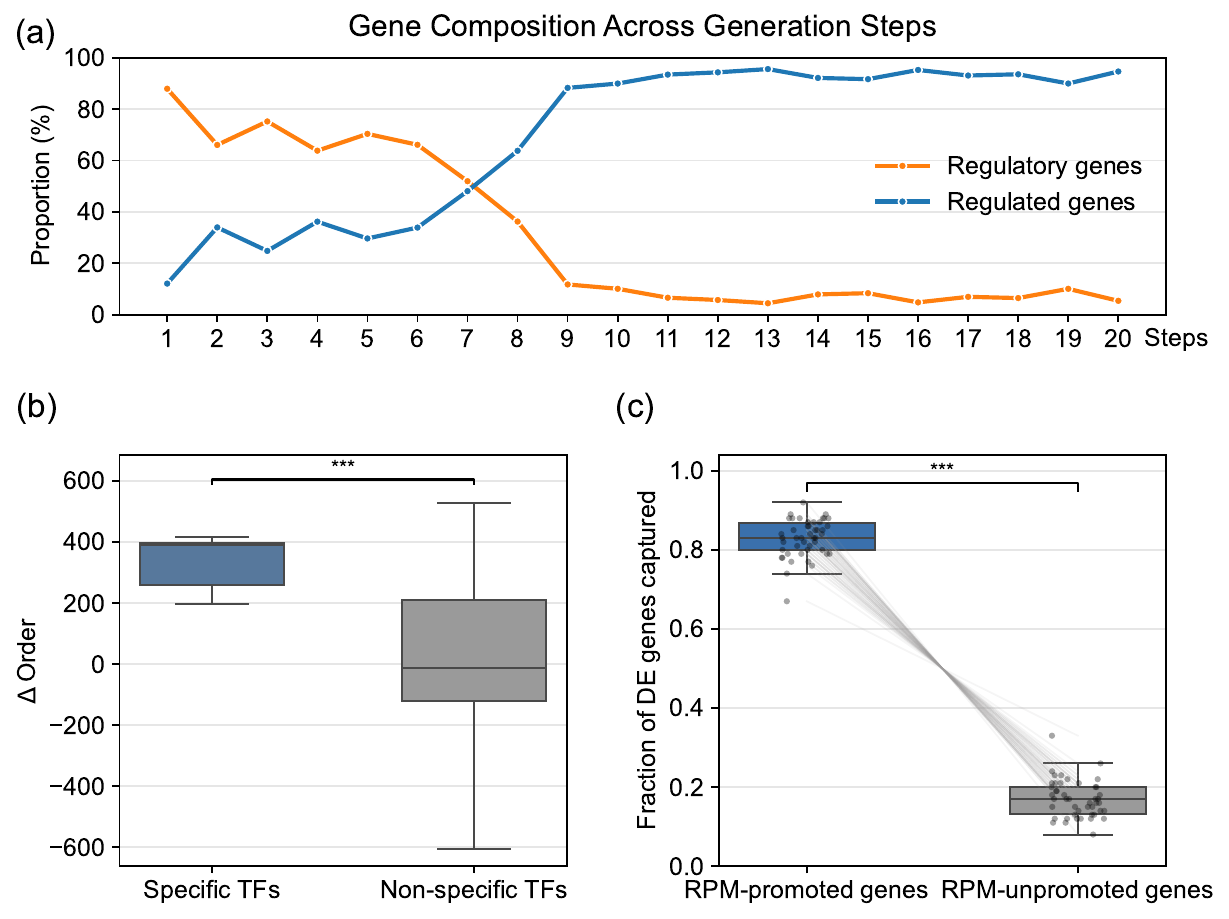}
    \caption{\textbf{Biological analysis.} (a) Proportions of regulatory and regulated genes across generation steps under the RPM-refined generation order. (b) Boxplot comparing $\Delta\mathrm{Order}$ between perturbation-specific TFs and non-specific TFs. Mann-Whitney $U$ test, $P = 1.6 \times 10^{-15}$. (c) Boxplot showing the fraction of DE genes whose generation order was promoted by RPM or remained unpromoted. Each dot represents one perturbation, and paired fractions from the same perturbation are connected by lines. Paired Wilcoxon test, $P = 1.7 \times 10^{-9}$.}
    \label{fig:fig4}
\end{figure}

Finally, we found that RPM-promoted genes (genes with $\Delta\mathrm{Order}>0$) were predominantly perturbation-responsive. As shown in Fig.~\ref{fig:fig4}c, RPM-promoted genes captured most perturbation-specific differentially expressed (DE) genes, and this pattern was consistent across all perturbations. RPM-promoted genes were also strongly enriched in cell-cycle-related pathways (see the Appendix), a major axis of perturbation response, especially given that many perturbations in Norman19 directly affect the proliferative state of cell. By prioritizing genes that characterize the perturbed cell state, RPM provides informative context for predicting the expression of later-generated genes.

\section{Conclusion}
We introduced $D^{2}R^{2}$, a framework that makes gene generation order explicit in single-cell perturbation prediction. MDDM progressively generates masked expression tokens, while RPM initializes the ordering policy from biological prior and adapts it to each perturbation through reinforcement learning. Across Norman19 and H1, $D^{2}R^{2}$ ranks first across all metrics on Norman19 and leads in most metrics on H1. Controlled ablations show that biologically informed ordering is more effective than random and uncertainty-based strategies. Biological analyses further reveal that RPM preserves upstream regulatory context while prioritizing perturbation-specific regulators and responsive genes.

\FloatBarrier
\putbib[refs]
\end{bibunit}

\appendix
\twocolumn[
\begin{center}
{\LARGE\bfseries Appendix}
\end{center}
\vspace{1em}
]
\begin{bibunit}[aaai2027]
\section{Background of Virtual Cell}
\label{app:virtual_cell_modeling}

Cell, the fundamental functional unit of life, is a complex entity. The intricate cellular behaviors emerge from highly coordinated interactions among genes, proteins, metabolites, and signaling pathways across multiple biological scales~\citep{yang2025build}. Understanding how these molecular systems collectively determine cellular responses to environmental stimuli and perturbations has long been a central goal in biology and medicine. However, the complexity, heterogeneity, and context dependency of cellular systems make comprehensive experimental characterization extremely challenging.

To better understand and predict cellular behaviors, researchers have increasingly turned toward the construction of virtual cell models to simulate, predict, and steer cell behavior~\citep{adduri2025predicting,bereket2023modelling,bunne2023learning,cui2024scgpt,he2026squidiff,klein2025cellflow}. The overarching vision of virtual cell modeling is to create predictive and mechanistic representations of cellular systems that can simulate how cells respond to perturbations, transition across developmental dynamics, and reorganize regulatory programs under varying biological conditions.

Bunne et al.~\citep{bunne2024build} proposed the definition of Artificial Intelligence Virtual Cell (AIVC), which is a multi-scale foundation model that learns universal representations of biological entities across molecular, cellular, and tissue scales. Within this framework, cellular systems are represented through learned embeddings derived from multi-modal biological measurements, including genomic sequences, transcriptomics, proteomics, imaging, and spatial omics data. These representations can then be manipulated by neural network-based ``virtual instruments'' that simulate cellular transitions, perturbation responses, and developmental dynamics. In essence, the virtual cell paradigm seeks to transform cellular biology into a programmable computational system capable of both predictive simulation and mechanistic inference.

The development of virtual cells has substantial scientific and clinical significance. Accurate virtual cell systems could enable large-scale in silico experimentation that would otherwise be prohibitively expensive or technically infeasible in physical laboratories. Furthermore, virtual cells may provide a unified computational framework for studying how molecular interactions collectively give rise to emergent cellular phenotypes across diverse biological contexts. Such systems have the potential to accelerate functional genomics, uncover regulatory mechanisms underlying development and disease, guide therapeutic target discovery, and improve precision medicine through patient-specific response prediction.

The rapid development of high-throughput experimental techniques has significantly advanced progress in virtual cell modeling. CRISPR-based perturb-seq~\citep{dixit2016perturb}, perturbation proteomics~\citep{qian2024ai}, and single-cell imaging~\citep{chandrasekaran2024three} now allow researchers to measure cellular states at unprecedented scale and resolution. In particular, datasets generated from genetic and chemical perturbation experiments~\citep{heumos2025pertpy} provide paired observations of pre- and post-perturbation cellular states, offering valuable resources for learning cellular response dynamics.

Driven by advances in experimental technologies, artificial intelligence has become a central methodology for virtual cell modeling. Given an initial cell state together with perturbation conditions, existing approaches aim to generate the corresponding post-perturbation cellular response. Current efforts mainly focus on perturbation-response prediction using transcriptomic measurements like scRNA-seq, since gene expression profiles provide scalable, information-rich representations of cellular states and are currently the most widely available modality for large-scale perturbation datasets. Existing approaches typically formulate virtual cell modeling as a conditional generative learning problem. Variational autoencoders~\citep{bereket2023modelling}, transformer architectures~\citep{adduri2025predicting}, graph neural networks~\citep{roohani2024predicting}, and more recently diffusion-based generative models~\citep{he2026squidiff} are commonly employed to learn the mappings between perturbations and transcriptomic responses. These models aim to capture the high-dimensional structure of cellular states while generalizing across unseen perturbations, cell types, and biological conditions.

Despite rapid progress, building reliable virtual cell systems remains an open challenge. Cellular systems exhibit complex nonlinear regulatory interactions, dynamic rewiring of gene regulatory networks, substantial biological heterogeneity, and strong context dependency across developmental and disease states~\citep{yang2025build}. Moreover, current models often prioritize predictive accuracy while lacking mechanistic interpretability and biologically grounded generation processes. Future virtual cell systems must not only generalize across modalities and perturbations, but also support mechanistic reasoning, uncertainty estimation, and biologically faithful simulation of cellular dynamics~\citep{wei2026benchmarking}. Developing deep learning architectures that can effectively integrate biological priors with generative modeling therefore remains a critical direction for future virtual cell research.

\section{Experimental Details}
\label{app:experimental_details}

\subsection{Datasets, Preprocessing, and Splits}

For both Norman19 and H1, we retain the top 1000 highly variable genes and keep
their identities fixed throughout tokenization and generation. Binning
statistics are fitted on training data and reused for validation and test
examples. Control cells provide the reference expression used to form
perturbation-induced changes.

Norman19 contains 287 perturbation conditions, including 131 combinatorial
conditions. We use the combination-prediction split with splitter seed 42;
held-out combinations are reserved for evaluation. H1 uses the benchmark's
saved held-out-cell split. The dataset split is fixed across repeated runs,
and the test split is used only for final evaluation.

\subsection{Evaluation Metrics}
\label{app:evaluation_metrics}

Following the PerturBench protocol cited in the main paper, metrics are computed for each
perturbation condition and then averaged across evaluation conditions. Let
$\hat{\mathbf y}_p$, $\mathbf y_p$, and $\mathbf y_{\mathrm{ctrl}}$ denote the
predicted, observed, and control mean-expression profiles for condition $p$.

\begin{itemize}
    \item \textbf{Pearson $\Delta$} is the Pearson correlation between
    $\hat{\mathbf y}_p-\mathbf y_{\mathrm{ctrl}}$ and
    $\mathbf y_p-\mathbf y_{\mathrm{ctrl}}$.
    \item \textbf{Cos. LogFC} is the cosine similarity between predicted and
    observed log$_2$ fold-change vectors relative to control, using a
    pseudocount of 0.1.
    \item \textbf{Cos. LogFC Rank} ranks the matched observed perturbation
    against the other conditions by LogFC cosine similarity. Lower values
    indicate better perturbation-level discrimination.
    \item \textbf{Cos. PCA} is the cosine similarity between predicted and
    observed condition centroids after projection into the reference PCA
    space.
    \item \textbf{Sym. KL} is the symmetrized KL divergence between predicted
    and observed cell populations in the same reference PCA space. Lower
    values indicate closer distributional agreement.
\end{itemize}

\subsection{Baseline Protocol}
\label{app:baseline_implementation}

\paragraph{Comparison protocol.}
We designed the comparison so that methods differ in model design rather than
in the data or evaluator. All methods use the same 1000-gene prediction space,
perturbation and control annotations, training/validation/test partitions, and
held-out test conditions. Predicted cells from every method are converted to
the same AnnData schema and evaluated by the same PerturBench-compatible code,
with the same reference controls, reference PCA space, condition aggregation,
and metric definitions. No method-specific metric implementation or test-set
post-processing is used.

We preserve model-specific input representations and training procedures from
the released implementations rather than forcing all methods into the discrete
representation used by $D^{2}R^{2}$. For methods integrated into the benchmark
(SAMS-VAE, Biolord, and Cell Flow), we use its unified data-loading and
training interfaces. For standalone methods, including STATE, Squidiff, and
scDFM, we adapt only the input/output interface required by the common split
and evaluator. Model selection and early stopping use validation behavior
only; test metrics are computed after model selection. Early stopping or a
fixed training horizon follows the corresponding implementation, because
an identical number of updates is not comparable across VAE, Transformer,
diffusion, and flow-based objectives.

\begin{table}[t]
\centering
\caption{Model-specific configurations used for the baseline methods.}
\label{tab:baseline_configuration}
\scriptsize
\setlength{\tabcolsep}{3pt}
\renewcommand{\arraystretch}{0.98}
\begin{tabular}{@{}p{0.18\columnwidth}p{0.76\columnwidth}@{}}
\toprule
Method & Configuration \\
\midrule
SAMS-VAE &
Latent size 128; batch 256; dropout 0.4; learning rate
$3.03\times10^{-5}$ \\

Biolord &
Latent size 512; batch 1000; dropout 0.4; learning rate
$1.67\times10^{-4}$ \\

STATE &
Hidden size 768; eight Transformer layers; 12 attention heads;
learning rate $10^{-4}$ \\

Cell Flow &
Batch 2000; hidden size 2048 with three layers; learning rate
$10^{-5}$ \\

Squidiff &
Three layers; batch 64; learning rate $10^{-4}$; EMA 0.9999 \\

scDFM &
Hidden size 512; batch 32; learning rate $10^{-5}$ \\
\bottomrule
\end{tabular}
\end{table}

\subsection{$D^{2}R^{2}$ Configuration}

The denoising predictor uses 50 expression bins and a 12-layer Transformer
with hidden dimension 768, 12 attention heads, and a SwiGLU intermediate
dimension of 3072. We use RMSNorm and attention and path dropout of 0.1.
Norman19 uses 64 control tokens and two perturbation
tokens, whereas H1 uses one control token and four perturbation tokens. At
inference time, generation proceeds for 20 steps and updates 50 genes per
step, with an MDDM sampling temperature of 1.0.

\paragraph{MDDM training.}
We first train the denoising predictor with the masked-token objective described
in the main paper. The optimization settings are summarized in
Table~\ref{tab:mddm_training_configuration}; the dataset-specific learning rate
is selected on the validation split.

\paragraph{GRPO training.}
After MDDM training, its parameters remain fixed while the RPM is optimized
with the GRPO objective defined in the main paper. Each group contains four
generation rollouts for the same example, and the resulting group-relative
advantages are used for one clipped policy update. The complete optimization
settings are given in Table~\ref{tab:grpo_training_configuration}.

\begin{table}[t]
\centering
\caption{MDDM training configuration. Dataset-specific values are
reported as Norman19 / H1.}
\label{tab:mddm_training_configuration}
\footnotesize
\setlength{\tabcolsep}{3pt}
\renewcommand{\arraystretch}{1.0}
\begin{tabular}{@{}p{0.51\columnwidth}p{0.41\columnwidth}@{}}
\toprule
Parameter & Value \\
\midrule
Batch size & 128 \\
Optimizer & AdamW \\
Weight decay & $10^{-2}$ \\
Learning rate (Norman19 / H1) & $10^{-4}$ / $5\times10^{-4}$ \\
Warmup fraction & 0.01 \\
Gradient clipping & 1.0 \\
Precision & bfloat16 mixed \\
EMA decay & 0.998 \\
\bottomrule
\end{tabular}
\end{table}

\begin{table}[t]
\centering
\caption{GRPO training configuration for RPM optimization.}
\label{tab:grpo_training_configuration}
\footnotesize
\setlength{\tabcolsep}{3pt}
\renewcommand{\arraystretch}{1.0}
\begin{tabular}{@{}p{0.51\columnwidth}p{0.41\columnwidth}@{}}
\toprule
Parameter & Value \\
\midrule
Batch size & 96 \\
Group size & 4 \\
Optimizer & AdamW \\
Learning rate & $3\times10^{-6}$ \\
Weight decay & $10^{-4}$ \\
Warmup fraction & 0.01 \\
Precision & bfloat16 mixed \\
Gradient clipping & 1.0 \\
GRPO clipping threshold & 0.15 \\
KL coefficient & 0.05 \\
Optimization iterations & 1 \\
Update cycles & 50 \\
\bottomrule
\end{tabular}
\end{table}

\paragraph{Selection of bin representative values.}
We explored two strategies for mapping generated discrete tokens back to
continuous expression values: assigning each sampled token the midpoint of its
corresponding bin, and taking the posterior expectation over all bin
midpoints. The expectation-based readout produced similar or slightly better
performance on mean-profile metrics, but substantially degraded
distribution-sensitive metrics such as MMD-PCA and Sym. KL, likely because
averaging across bins reduces the variability of the generated cell
population. We therefore use the bin-midpoint readout throughout our
experiments. This mapping is deterministic and introduces no additional
trainable parameters.

\subsection{Repeated Runs and Statistical Protocol}

The split is fixed across runs. The five reported seeds are
$\{0,1,2,3,42\}$; they control training randomness, including initialization,
data shuffling, and stochastic optimization, rather than dataset construction.
Results in the main-results table are mean $\pm$ sample standard
deviation over five independent runs. For $D^{2}R^{2}$, the MDDM checkpoint is
fixed and the adaptive RPM branch is independently initialized and optimized
for each seed. The generation-order ablation likewise fixes the MDDM and the
20-step generation budget so that only the selection policy changes.

Lower-is-better metrics are
sign-reversed when reporting paired improvements so that a positive difference
always denotes improvement. These condition-level intervals are distinct from
the training-seed standard deviations in the main table.

\subsection{Compute Environment}

Experiments were run under Ubuntu 22.04 with Python 3.10, CUDA 12.4,
PyTorch 2.6, Lightning 2.6, and Scanpy 1.11 on NVIDIA A100 80GB PCIe GPUs.
MDDM training and RPM optimization use bfloat16 mixed precision. Metric
computation uses the PerturBench-compatible evaluation implementation released
with the code.

\section{Biological Analysis}
\label{app:biological_analysis}

\subsection{Analysis of regulatory-gene composition across generation steps.}

We inferred a cell-type-specific gene regulatory network (GRN) from unperturbed control cells using DeepSEM~\citep{shu2021modeling}. We then assigned the 1,000 genes generated by $D^{2}R^{2}$ to two mutually exclusive categories: regulatory genes, defined as genes with at least one outgoing edge in the GRN, and the remaining genes are defined as regulated genes. This classification yielded 303 regulatory genes and 697 regulated genes. For each generation step, we calculated the proportion of generated genes in each category and plotted these proportions across steps.

\subsection{Comparison of $\Delta\mathrm{Order}$ between perturbation-specific and non-specific TFs.}

In this analysis, we identified perturbation-specific transcription factors (TFs) from DoRothEA regulon activity~\citep{garcia2019benchmark} using perturbation-versus-control expression changes. DoRothEA regulons were restricted to TF-target pairs represented among the genes modeled by $D^{2}R^{2}$. For each perturbation, we calculated the z-scored expression change of each gene and tested whether the changes across each TF’s target genes deviated significantly from the background distribution. The resulting $P$ values were adjusted using the Benjamini–Hochberg procedure. TFs with an $\mathrm{FDR} < 0.1$ were classified as perturbation-specific, whereas all remaining tested TFs were classified as non-specific. For each TF, we calculated $\Delta\mathrm{Order}$ as its biological-prior order minus its RPM-refined order. Thus, $\Delta\mathrm{Order}>0$ indicates that RPM moved the TF to an earlier generation step. We compared $\Delta\mathrm{Order}$ between the two groups using a one-sided Mann–Whitney $U$ test.

\begin{figure*}[t]
    \centering
    \includegraphics[width=\textwidth]{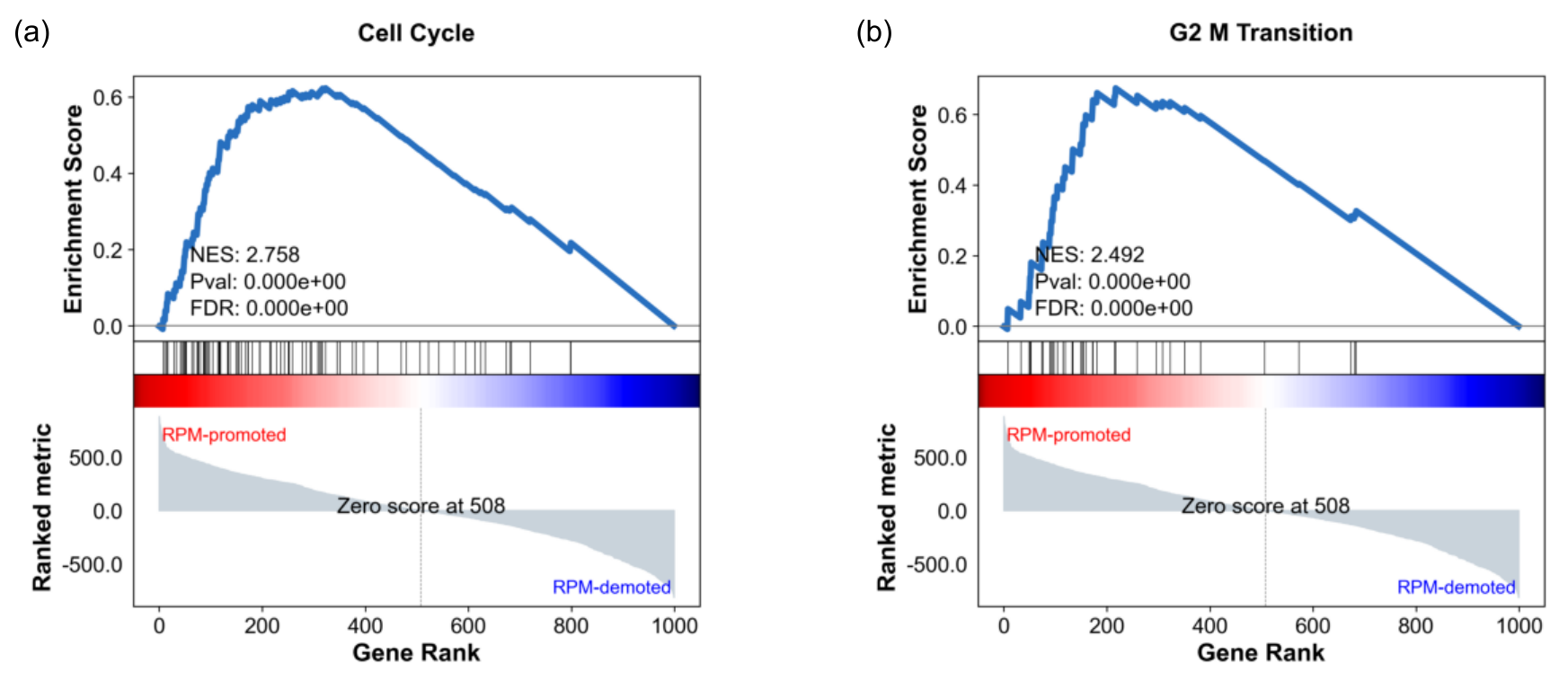}
    \caption{\textbf{GSEA analysis using the $\Delta\mathrm{Order}$ values of all genes as the ranking metric.} The GSEA analysis shows that RPM-promoted genes are enriched in cell cycle pathway (a) and G2-M transition pathway (b).}
    \label{fig:fig5}
\end{figure*}

\subsection{Analysis of $\Delta\mathrm{Order}$ among differentially expressed genes.}

Genes with $\Delta\mathrm{Order}>0$ were defined as RPM-promoted genes, whereas those with $\Delta\mathrm{Order}\leq0$ were defined as RPM-unpromoted genes. For each perturbation, we performed differential expression analysis against unperturbed control cells using \texttt{scanpy.tl.rank\_genes\_groups}. The top 100 genes with the largest absolute differential expression scores were defined as DE genes for that perturbation. Then, we calculated the fractions of DE genes classified as RPM-promoted and unpromoted, and compared the paired fractions across perturbations using a one-sided Wilcoxon signed-rank test.

\subsection{Gene set enrichment analysis (GSEA) of $\Delta\mathrm{Order}$.}

Gene set enrichment analysis was performed on the averaged $\Delta\mathrm{Order}$ across perturbations using \texttt{gseapy.prerank} with Reactome Pathways 2024. A positive normalized enrichment score (NES) indicates enrichment among genes with higher $\Delta\mathrm{Order}$, corresponding to genes promoted earlier by RPM.

\section{The Results of GSEA Analysis}

As shown in Fig.~\ref{fig:fig5}, the RPM-promoted genes are strongly enriched in cell-cycle-related pathways, a major axis of perturbation response.

\FloatBarrier
\putbib[refs]
\end{bibunit}


\begin{thebibliography}{36}
\providecommand{\natexlab}[1]{#1}

\bibitem[{Adduri et~al.(2025)Adduri, Gautam, Bevilacqua, Imran, Shah,
  Naghipourfar, Teyssier, Ilango, Nagaraj, Dong et~al.}]{adduri2025predicting}
Adduri, A.~K.; Gautam, D.; Bevilacqua, B.; Imran, A.; Shah, R.; Naghipourfar,
  M.; Teyssier, N.; Ilango, R.; Nagaraj, S.; Dong, M.; et~al. 2025.
\newblock Predicting cellular responses to perturbation across diverse contexts
  with State.
\newblock \emph{BioRxiv}, 2025--06.

\bibitem[{Austin et~al.(2021)Austin, Johnson, Ho, Tarlow, and Van
  Den~Berg}]{austin2021structured}
Austin, J.; Johnson, D.~D.; Ho, J.; Tarlow, D.; and Van Den~Berg, R. 2021.
\newblock Structured denoising diffusion models in discrete state-spaces.
\newblock \emph{Advances in neural information processing systems}, 34:
  17981--17993.

\bibitem[{Bahdanau et~al.(2016)Bahdanau, Brakel, Xu, Goyal, Lowe, Pineau,
  Courville, and Bengio}]{bahdanau2016actor}
Bahdanau, D.; Brakel, P.; Xu, K.; Goyal, A.; Lowe, R.; Pineau, J.; Courville,
  A.; and Bengio, Y. 2016.
\newblock An actor-critic algorithm for sequence prediction.
\newblock \emph{arXiv preprint arXiv:1607.07086}.

\bibitem[{Bereket and Karaletsos(2023)}]{bereket2023modelling}
Bereket, M.; and Karaletsos, T. 2023.
\newblock Modelling cellular perturbations with the sparse additive mechanism
  shift variational autoencoder.
\newblock \emph{Advances in Neural Information Processing Systems}, 36: 1--12.

\bibitem[{Bunne et~al.(2024)Bunne, Roohani, Rosen, Gupta, Zhang, Roed,
  Alexandrov, AlQuraishi, Brennan, Burkhardt et~al.}]{bunne2024build}
Bunne, C.; Roohani, Y.; Rosen, Y.; Gupta, A.; Zhang, X.; Roed, M.; Alexandrov,
  T.; AlQuraishi, M.; Brennan, P.; Burkhardt, D.~B.; et~al. 2024.
\newblock How to build the virtual cell with artificial intelligence:
  Priorities and opportunities.
\newblock \emph{Cell}, 187(25): 7045--7063.

\bibitem[{Bunne et~al.(2023)Bunne, Stark, Gut, Del~Castillo, Levesque, Lehmann,
  Pelkmans, Krause, and R{\"a}tsch}]{bunne2023learning}
Bunne, C.; Stark, S.~G.; Gut, G.; Del~Castillo, J.~S.; Levesque, M.; Lehmann,
  K.-V.; Pelkmans, L.; Krause, A.; and R{\"a}tsch, G. 2023.
\newblock Learning single-cell perturbation responses using neural optimal
  transport.
\newblock \emph{Nature methods}, 20(11): 1759--1768.

\bibitem[{Chang et~al.(2022)Chang, Zhang, Jiang, Liu, and
  Freeman}]{chang2022maskgit}
Chang, H.; Zhang, H.; Jiang, L.; Liu, C.; and Freeman, W.~T. 2022.
\newblock Maskgit: Masked generative image transformer.
\newblock In \emph{Proceedings of the IEEE/CVF conference on computer vision
  and pattern recognition}, 11315--11325.

\bibitem[{Cui et~al.(2024{\natexlab{a}})Cui, Wang, Maan, Pang, Luo, Duan, and
  Wang}]{cui2024scgpt}
Cui, H.; Wang, C.; Maan, H.; Pang, K.; Luo, F.; Duan, N.; and Wang, B.
  2024{\natexlab{a}}.
\newblock scGPT: toward building a foundation model for single-cell multi-omics
  using generative AI.
\newblock \emph{Nature methods}, 21(8): 1470--1480.

\bibitem[{Cui et~al.(2024{\natexlab{b}})Cui, Xu, Wang, Liao, and
  Wang}]{cui2024geneformer}
Cui, Z.; Xu, T.; Wang, J.; Liao, Y.; and Wang, Y. 2024{\natexlab{b}}.
\newblock Geneformer: Learned gene compression using transformer-based context
  modeling.
\newblock In \emph{ICASSP 2024-2024 IEEE International Conference on Acoustics,
  Speech and Signal Processing (ICASSP)}, 8035--8039. IEEE.

\bibitem[{Dixit et~al.(2016)Dixit, Parnas, Li, Chen, Fulco, Jerby-Arnon,
  Marjanovic, Dionne, Burks, Raychowdhury et~al.}]{dixit2016perturb}
Dixit, A.; Parnas, O.; Li, B.; Chen, J.; Fulco, C.~P.; Jerby-Arnon, L.;
  Marjanovic, N.~D.; Dionne, D.; Burks, T.; Raychowdhury, R.; et~al. 2016.
\newblock Perturb-Seq: dissecting molecular circuits with scalable single-cell
  RNA profiling of pooled genetic screens.
\newblock \emph{cell}, 167(7): 1853--1866.

\bibitem[{Freimer et~al.(2022)Freimer, Shaked, Naqvi, Sinnott-Armstrong,
  Kathiria, Garrido, Chen, Cortez, Greenleaf, Pritchard
  et~al.}]{freimer2022systematic}
Freimer, J.~W.; Shaked, O.; Naqvi, S.; Sinnott-Armstrong, N.; Kathiria, A.;
  Garrido, C.~M.; Chen, A.~F.; Cortez, J.~T.; Greenleaf, W.~J.; Pritchard,
  J.~K.; et~al. 2022.
\newblock Systematic discovery and perturbation of regulatory genes in human T
  cells reveals the architecture of immune networks.
\newblock \emph{Nature Genetics}, 54(8): 1133--1144.

\bibitem[{He et~al.(2026)He, Zhu, Tavakol, Ye, Lao, Zhu, Xu, Chauhan, Garty,
  Tomer et~al.}]{he2026squidiff}
He, S.; Zhu, Y.; Tavakol, D.~N.; Ye, H.; Lao, Y.-H.; Zhu, Z.; Xu, C.; Chauhan,
  S.; Garty, G.; Tomer, R.; et~al. 2026.
\newblock Squidiff: predicting cellular development and responses to
  perturbations using a diffusion model.
\newblock \emph{Nature methods}, 23(1): 65--77.

\bibitem[{Klein et~al.(2025)Klein, Fleck, Bobrovskiy, Zimmermann, Becker,
  Palma, Dony, Tejada-Lapuerta, Huguet, Lin et~al.}]{klein2025cellflow}
Klein, D.; Fleck, J.~S.; Bobrovskiy, D.; Zimmermann, L.; Becker, S.; Palma, A.;
  Dony, L.; Tejada-Lapuerta, A.; Huguet, G.; Lin, H.-C.; et~al. 2025.
\newblock CellFlow enables generative single-cell phenotype modeling with flow
  matching.
\newblock \emph{bioRxiv}, 2025--04.

\bibitem[{Lotfollahi, Wolf, and Theis(2019)}]{lotfollahi2019scgen}
Lotfollahi, M.; Wolf, F.~A.; and Theis, F.~J. 2019.
\newblock scGen predicts single-cell perturbation responses.
\newblock \emph{Nature methods}, 16(8): 715--721.

\bibitem[{Lutz et~al.(2023)Lutz, Wang, Norn, Courbet, Borst, Zhao, Dosey, Cao,
  Xu, Leaf et~al.}]{lutz2023top}
Lutz, I.~D.; Wang, S.; Norn, C.; Courbet, A.; Borst, A.~J.; Zhao, Y.~T.; Dosey,
  A.; Cao, L.; Xu, J.; Leaf, E.~M.; et~al. 2023.
\newblock Top-down design of protein architectures with reinforcement learning.
\newblock \emph{Science}, 380(6642): 266--273.

\bibitem[{Nie et~al.(2025{\natexlab{a}})Nie, Zhu, Du, Pang, Liu, Zeng, Lin, and
  Li}]{nie2025scaling}
Nie, S.; Zhu, F.; Du, C.; Pang, T.; Liu, Q.; Zeng, G.; Lin, M.; and Li, C.
  2025{\natexlab{a}}.
\newblock Scaling up Masked Diffusion Models on Text.
\newblock In \emph{The Thirteenth International Conference on Learning
  Representations}.

\bibitem[{Nie et~al.(2025{\natexlab{b}})Nie, Zhu, You, Zhang, Ou, Hu, Zhou,
  Lin, Wen, and Li}]{nie2025large}
Nie, S.; Zhu, F.; You, Z.; Zhang, X.; Ou, J.; Hu, J.; Zhou, J.; Lin, Y.; Wen,
  J.-R.; and Li, C. 2025{\natexlab{b}}.
\newblock Large language diffusion models.
\newblock \emph{arXiv preprint arXiv:2502.09992}.

\bibitem[{Norman et~al.(2019)Norman, Horlbeck, Replogle, Ge, Xu, Jost, Gilbert,
  and Weissman}]{norman2019exploring}
Norman, T.~M.; Horlbeck, M.~A.; Replogle, J.~M.; Ge, A.~Y.; Xu, A.; Jost, M.;
  Gilbert, L.~A.; and Weissman, J.~S. 2019.
\newblock Exploring genetic interaction manifolds constructed from rich
  single-cell phenotypes.
\newblock \emph{Science}, 365(6455): 786--793.

\bibitem[{Olivecrona et~al.(2017)Olivecrona, Blaschke, Engkvist, and
  Chen}]{olivecrona2017molecular}
Olivecrona, M.; Blaschke, T.; Engkvist, O.; and Chen, H. 2017.
\newblock Molecular de-novo design through deep reinforcement learning.
\newblock \emph{Journal of cheminformatics}, 9(1): 48.

\bibitem[{Piran et~al.(2024)Piran, Cohen, Hoshen, and
  Nitzan}]{piran2024disentanglement}
Piran, Z.; Cohen, N.; Hoshen, Y.; and Nitzan, M. 2024.
\newblock Disentanglement of single-cell data with biolord.
\newblock \emph{Nature Biotechnology}, 42(11): 1678--1683.

\bibitem[{Qian, Dong, and Guo(2025)}]{qian2025grow}
Qian, L.; Dong, Z.; and Guo, T. 2025.
\newblock Grow AI virtual cells: three data pillars and closed-loop learning.
\newblock \emph{Cell Research}, 35(5): 319--321.

\bibitem[{Ranzato et~al.(2015)Ranzato, Chopra, Auli, and
  Zaremba}]{ranzato2015sequence}
Ranzato, M.; Chopra, S.; Auli, M.; and Zaremba, W. 2015.
\newblock Sequence level training with recurrent neural networks.
\newblock \emph{arXiv preprint arXiv:1511.06732}.

\bibitem[{Roohani, Huang, and Leskovec(2024)}]{roohani2024predicting}
Roohani, Y.; Huang, K.; and Leskovec, J. 2024.
\newblock Predicting transcriptional outcomes of novel multigene perturbations
  with GEARS.
\newblock \emph{Nature Biotechnology}, 42(6): 927--935.

\bibitem[{Roohani et~al.(2025)Roohani, Hua, Tung, Bounds, Yu, Dobin, Teyssier,
  Adduri, Woodrow, Plosky et~al.}]{roohani2025virtual}
Roohani, Y.~H.; Hua, T.~J.; Tung, P.-Y.; Bounds, L.~R.; Yu, F.~B.; Dobin, A.;
  Teyssier, N.; Adduri, A.; Woodrow, A.; Plosky, B.~S.; et~al. 2025.
\newblock Virtual Cell Challenge: Toward a Turing test for the virtual cell.
\newblock \emph{Cell}, 188(13): 3370--3374.

\bibitem[{Runge et~al.(2018)Runge, Stoll, Falkner, and
  Hutter}]{runge2018learning}
Runge, F.; Stoll, D.; Falkner, S.; and Hutter, F. 2018.
\newblock Learning to design RNA.
\newblock \emph{arXiv preprint arXiv:1812.11951}.

\bibitem[{Settles(2009)}]{settles2009active}
Settles, B. 2009.
\newblock Active learning literature survey.

\bibitem[{Shao et~al.(2024)Shao, Wang, Zhu, Xu, Song, Bi, Zhang, Zhang, Li, Wu
  et~al.}]{shao2024deepseekmath}
Shao, Z.; Wang, P.; Zhu, Q.; Xu, R.; Song, J.; Bi, X.; Zhang, H.; Zhang, M.;
  Li, Y.; Wu, Y.; et~al. 2024.
\newblock Deepseekmath: Pushing the limits of mathematical reasoning in open
  language models.
\newblock \emph{arXiv preprint arXiv:2402.03300}.

\bibitem[{Shu et~al.(2021)Shu, Zhou, Lian, Li, Zhao, Zeng, and
  Ma}]{shu2021modeling}
Shu, H.; Zhou, J.; Lian, Q.; Li, H.; Zhao, D.; Zeng, J.; and Ma, J. 2021.
\newblock Modeling gene regulatory networks using neural network architectures.
\newblock \emph{Nature Computational Science}, 1(7): 491--501.

\bibitem[{Song et~al.(2025)Song, Liu, Dai, Mcmyn, Wang, Yang, Krejci, Vasilyev,
  Untermoser, Loregger et~al.}]{song2025decoding}
Song, B.; Liu, D.; Dai, W.; Mcmyn, N.~F.; Wang, Q.; Yang, D.; Krejci, A.;
  Vasilyev, A.; Untermoser, N.; Loregger, A.; et~al. 2025.
\newblock Decoding heterogeneous single-cell perturbation responses.
\newblock \emph{Nature cell biology}, 27(3): 493--504.

\bibitem[{Tang et~al.(2023)Tang, Liu, Wen, Dai, Ding, Li, Fan, Xie, and
  Tang}]{tang2023general}
Tang, W.; Liu, R.; Wen, H.; Dai, X.; Ding, J.; Li, H.; Fan, W.; Xie, Y.; and
  Tang, J. 2023.
\newblock A general single-cell analysis framework via conditional diffusion
  generative models.
\newblock \emph{bioRxiv}, 2023--10.

\bibitem[{Theodoris et~al.(2023)Theodoris, Xiao, Chopra, Chaffin, Al~Sayed,
  Hill, Mantineo, Brydon, Zeng, Liu et~al.}]{theodoris2023transfer}
Theodoris, C.~V.; Xiao, L.; Chopra, A.; Chaffin, M.~D.; Al~Sayed, Z.~R.; Hill,
  M.~C.; Mantineo, H.; Brydon, E.~M.; Zeng, Z.; Liu, X.~S.; et~al. 2023.
\newblock Transfer learning enables predictions in network biology.
\newblock \emph{Nature}, 618(7965): 616--624.

\bibitem[{Vaswani et~al.(2017)Vaswani, Shazeer, Parmar, Uszkoreit, Jones,
  Gomez, Kaiser, and Polosukhin}]{vaswani2017attention}
Vaswani, A.; Shazeer, N.; Parmar, N.; Uszkoreit, J.; Jones, L.; Gomez, A.~N.;
  Kaiser, {\L}.; and Polosukhin, I. 2017.
\newblock Attention is all you need.
\newblock \emph{Advances in neural information processing systems}, 30.

\bibitem[{Wu et~al.(2024)Wu, Wershof, Schmon, Nassar, Osi{\'n}ski, Eksi, Yan,
  Stark, Zhang, and Graepel}]{wu2024perturbench}
Wu, Y.; Wershof, E.; Schmon, S.~M.; Nassar, M.; Osi{\'n}ski, B.; Eksi, R.; Yan,
  Z.; Stark, R.; Zhang, K.; and Graepel, T. 2024.
\newblock Perturbench: Benchmarking machine learning models for cellular
  perturbation analysis.
\newblock \emph{arXiv preprint arXiv:2408.10609}.

\bibitem[{You et~al.(2018)You, Liu, Ying, Pande, and Leskovec}]{you2018graph}
You, J.; Liu, B.; Ying, Z.; Pande, V.; and Leskovec, J. 2018.
\newblock Graph convolutional policy network for goal-directed molecular graph
  generation.
\newblock \emph{Advances in neural information processing systems}, 31.

\bibitem[{Yu et~al.(2026)Yu, Wang, Liao, and Wu}]{yu2026scdfm}
Yu, C.; Wang, C.; Liao, B.; and Wu, T. 2026.
\newblock scdfm: Distributional flow matching model for robust single-cell
  perturbation prediction.
\newblock \emph{arXiv preprint arXiv:2602.07103}.

\bibitem[{Yuan et~al.(2021)Yuan, Shen, Luna, Korkut, Marks, Ingraham, and
  Sander}]{yuan2021cellbox}
Yuan, B.; Shen, C.; Luna, A.; Korkut, A.; Marks, D.~S.; Ingraham, J.; and
  Sander, C. 2021.
\newblock CellBox: interpretable machine learning for perturbation biology with
  application to the design of cancer combination therapy.
\newblock \emph{Cell systems}, 12(2): 128--140.

\end{thebibliography}
\end{document}